\documentclass[conference,a4paper]{IEEEtran}
\usepackage[T1]{fontenc}
\usepackage{amsmath,amssymb}
\usepackage{graphicx,booktabs,xcolor,float}
\usepackage{tikz,pgfplots}
\usetikzlibrary{arrows.meta,positioning,calc}
\usepgfplotslibrary{groupplots}
\pgfplotsset{compat=1.18}
\usepackage{cite}
\usepackage[hidelinks]{hyperref}
\usepackage{tabularx,array}
\newcommand{\Records}{5,183}
\newcommand{\SixSpots}{4,883}
\newcommand{\SixObs}{5,553}
\newcommand{\SixMerged}{670}

\newcommand{\KuhnIters}{10,000}
\newcommand{\KuhnTeacherExpl}{0.001486}
\newcommand{\LeducIters}{2,000}
\newcommand{\LeducTeacherExpl}{0.010877}
\newcommand{\SplitKuhn}{6/3/3}
\newcommand{\SplitLeduc}{189/45/54}
\newcommand{\SplitSix}{3,449/705/729}

\newcommand{\FeatDim}{4,096}

\hypersetup{pdftitle={Teaching AI Agent to Play Poker Strategically Through Preference Optimization},pdfauthor={Phurinat Polasa, Thanapon Thanadunpremdet, Phatthanasak Kraiduang, Saksorn Ruangtanusak}}

\makeatletter
\newcommand{\linebreakand}{%
  \end{@IEEEauthorhalign}
  \hfill\mbox{}\par
  \mbox{}\hfill\begin{@IEEEauthorhalign}
}
\makeatother

\newif\ifanonymous
\anonymousfalse 
\newcommand{\E}{\mathbb{E}}
\newcommand{\KL}{D_{\mathrm{KL}}}
\definecolor{spokeblue}{HTML}{245875}
\IEEEoverridecommandlockouts

\title{Teaching AI Agent to Play Poker Strategically Through Preference Optimization%
\thanks{Accepted at the iSAI-NLP 2026 Student Research Workshop (SRW).}}
\ifanonymous
\author{
\IEEEauthorblockN{[Anonymous Authors]}
\IEEEauthorblockA{[Institution, City, Country] \\ \quad [Email]}
}
\else
\author{
\IEEEauthorblockN{Phurinat Polasa}
\IEEEauthorblockA{
Department of Computer Engineering\\
KMUTT\\
Thailand\\
phurinat.pola@kmutt.ac.th}
\and
\IEEEauthorblockN{Thanapon Thanadunpremdet}
\IEEEauthorblockA{
Department of Computer Engineering\\
Mahidol University\\
Thailand\\
thanapon.tha@student.mahidol.edu}
\linebreakand
\IEEEauthorblockN{Phatthanasak Kraiduang}
\IEEEauthorblockA{
Department of Computer Engineering\\
Mahidol University\\
Thailand\\
phatthanasak.kra@student.mahidol.ac.th}
\and
\IEEEauthorblockN{Saksorn Ruangtanusak}
\IEEEauthorblockA{
R\&D\\
SCB DataX, SCBX Group\\
Thailand\\
saksorn.ruangtanusak@data-x.ai}
}
\fi

\begin{document}
\maketitle

\begin{abstract}
A poker agent can make a legal move and still make a strategically poor one. We ask whether compact AI agents can exhibit strategic behavior under hidden information and whether preference optimization can improve beyond static and zero-shot policies. We first evaluate a static baseline and three frozen language models: SmallThinker, Qwen3, and Gemma-3 whose performance varies substantially between exact-game and six-player evaluation. We therefore propose Strategic Preference Optimization with KL-anchored Equilibrium (SPOKE), combining teacher-relative value penalties, policy anchoring, risk constraints, and structured rationale supervision, and compare it with SFT, DPO, GRPO, and GSPO. Across three runs of a SmallThinker-4B-A0.6B policy, SPOKE achieves the lowest mean exploitability in Kuhn (0.213) and Leduc (1.081) and the highest six-max top-1 agreement (0.775), while maintaining teacher KL comparable to SFT (0.522 vs.\ 0.523). These results provide preliminary evidence that strategic preference optimization can improve strategic quality while preserving teacher-consistent behavior.
\end{abstract}


\section{Introduction}

Legal poker actions can still be strategically poor, making poker a useful testbed for strategic decision-making under hidden information. CFR provides approximate equilibrium strategies in finite two-player zero-sum games~\cite{cfr}, while systems such as Pluribus~\cite{pluribus} motivate evaluation in more complex multiplayer settings.

We first compare a static policy with three frozen compact language models: SmallThinker, Qwen3, and Gemma-3. SmallThinker performs best among them in exact games, yet remains weak on six-max agreement.

We therefore compare Supervised Fine-Tuning (SFT), Direct Preference Optimization (DPO)~\cite{dpo}, Group Relative Policy Optimization (GRPO)~\cite{grpo}, and Group Sequence Policy Optimization (GSPO)~\cite{gspo}, and propose Strategic Preference Optimization with KL-anchored Equilibrium (SPOKE). SPOKE combines teacher-relative value penalties, KL anchoring, risk constraints, and structured rationale supervision to improve strategic quality without sacrificing teacher-consistent behavior.

\begin{figure}
    \centering
    \includegraphics[width=0.95\linewidth]{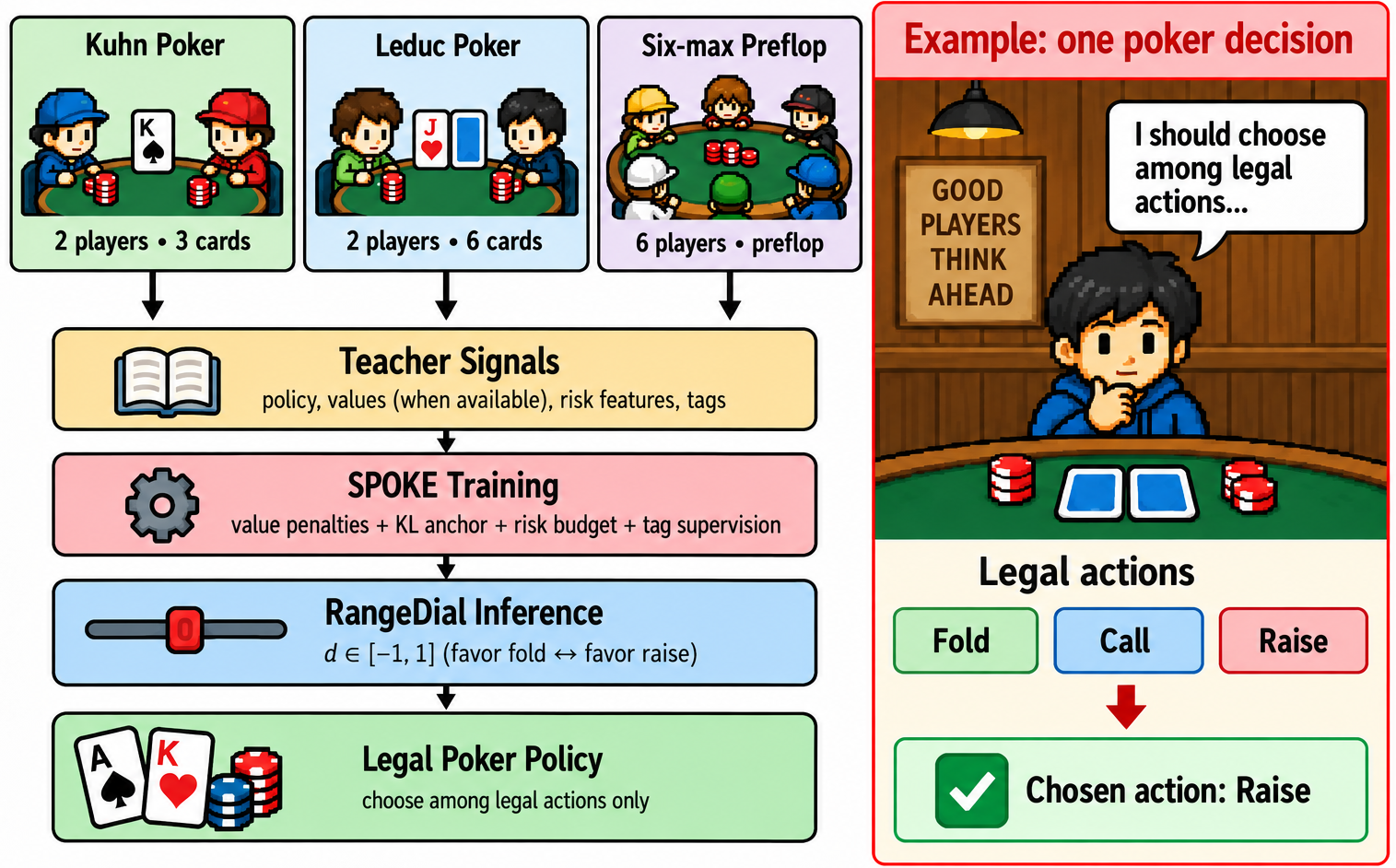}
    \caption{SPOKE overview: teaching a poker agent to act legally and strategically.}
    \label{fig:spoke_overview}
\end{figure}

\section{Proposed Method}
\begin{figure}[h]
\centering
\resizebox{0.9\columnwidth}{!}{
\begin{tikzpicture}[font=\scriptsize, node distance=2.4mm and 3mm,
  box/.style={draw, rounded corners=1.2pt, align=center, inner sep=2.5pt, minimum height=6mm},
  arr/.style={-{Latex[length=1.5mm]}, thick}]
\node[box, text width=30mm] (cfr) {Kuhn / limit Leduc\\CFR: $\pi_T,\,Q_T,\,V_T$};
\node[box, text width=30mm, right=6mm of cfr] (six) {Six-max preflop\\Smoothed behavior labels};
\node[box, text width=58mm, below=of $(cfr.south)!0.5!(six.south)$] (rec) {Decision records + grouped splits\\Legal menu, risk features, rationale tags};
\node[box, fill=spokeblue!14, draw=spokeblue, text width=28mm, below=of rec.south west, anchor=north west] (train) {SPOKE training\\Values + KL + risk + tags};
\node[box, text width=25mm, right=6mm of train] (filter) {Legal-action filtering\\RangeDial $d\in[-1,1]$};
\node[box, text width=58mm, below=of $(train.south)!0.5!(filter.south)$] (eval) {Separate evaluation + audit trail\\Exact-game exploitability / six-max behavior};
\draw[arr] (cfr.south) -- (cfr.south |- rec.north);
\draw[arr] (six.south) -- (six.south |- rec.north);
\draw[arr] (train.north |- rec.south) -- (train.north);
\draw[arr] (train) -- node[above, font=\tiny, inner sep=1pt] {policy} (filter);
\draw[arr] (filter.south) -- (filter.south |- eval.north);
\end{tikzpicture}}
\caption{Proposed SPOKE pipeline. Teacher values are used only where validated; six-max action labels provide behavioral supervision. Legal-action filtering is shared by all methods.}
\label{fig:method}
\end{figure}
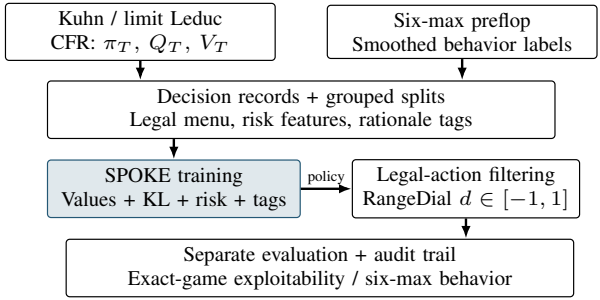

Each record contains the player's information state $s$, legal action--size menu $A(s)$, teacher distribution $\pi_T$, optional action values $Q_T$, risk features, and rationale tags $t$. Prompts exclude opponents' hidden cards. Canonical action strings receive model likelihood scores, normalized over $A(s)$ to form $\pi_\theta$; a separate output predicts structured tags (Fig.~\ref{fig:method}).

We operationalize the objective as follows. Let $[x]_+=\max(x,0)$ and $V_T=\E_{a\sim\pi_T}Q_T(s,a)$. Define
\begin{align}
R_s^+&=\E_{a\sim\pi_\theta}[V_T-Q_T(s,a)]_+,\nonumber\\
\Delta_s^+&=[V_T-\E_{a\sim\pi_\theta}Q_T(s,a)]_+ .\label{eq:value}
\end{align}
The first penalty discourages individual below-reference actions; the second penalizes an average value shortfall. These are teacher-relative proxies, not cumulative CFR regret. With $m_s$ indicating validated value targets, we optimize the policy using
\begin{align}
\mathcal{L}=\E_s\big[&m_s(\lambda_r R_s^++\lambda_v\Delta_s^+)
 +\lambda_k\KL(\pi_T\Vert\pi_\theta)\nonumber\\
 &+\lambda_b[\E_{a\sim\pi_\theta}c(s,a)-B_s]_+
 +\lambda_t\ell_{\mathrm{tag}}\big].\label{eq:loss}
\end{align}

Here $c$ is a risk cost, $B_s$ its budget, and $\ell_{\mathrm{tag}}$ is binary cross-entropy over rationale tags. KL anchoring discourages collapse of the teacher's mixed policy. We use value-based terms only where validated ($m_s=1$ for Kuhn/Leduc and $m_s=0$ for six-max), with hyperparameters selected on the development split.

After restricting the policy to legal actions, RangeDial applies a bounded preference shift
\begin{equation}
\pi_{\theta,d}(a\mid s)\propto\pi_\theta(a\mid s)\exp(\alpha d\,b(a)),\quad d\in[-1,1],
\end{equation}
where $b$ is $1$ for raise, $0$ for check/call, and $-1$ for fold; $\alpha\geq0$ sets the strength. Increasing $d$ favors raising without adding illegal actions.

\section{Experimental Setups}
\textit{Data and evaluation.} The record set holds \Records\ decisions after merging \SixMerged\ repeated six-max spots (\SixSpots\ spots from \SixObs\ observations), with teacher exploitability \KuhnTeacherExpl\ in Kuhn (\KuhnIters\ CFR iterations) and \LeducTeacherExpl\ in limit Leduc (\LeducIters\ iterations). Splits precede prompt augmentation and group information sets and public histories to prevent leakage: train/dev/test are \SplitKuhn\ (Kuhn), \SplitLeduc\ (Leduc) and \SplitSix\ (six-max). Full chance enumeration and best responses support exact profile value, NashConv, and exploitability on the declared small games (over all information sets).

Six-max evaluation uses 100-big-blind preflop decisions with label-smoothed, Pluribus-derived behavioral actions. The primary behavioral endpoint is held-out top-1 agreement; teacher KL is secondary. Legality is checked separately.

\textit{Models and controls.}
The static baseline samples uniformly from legal actions in Kuhn and Leduc and uses a fixed preflop action chart in six-max. Frozen SmallThinker-4B-A0.6B~\cite{smallthinker}, Qwen3-0.6B~\cite{qwen}, and Gemma-3-1B-IT~\cite{gemma} were each evaluated once on the same held-out sets, task prompt, and legal menu. Action probabilities use first-token scores for \emph{fold}/\emph{call}/\emph{raise}; these serve as zero-shot controls.

We then train SFT, DPO~\cite{dpo}, GRPO~\cite{grpo}, GSPO~\cite{gspo}, and SPOKE as a SmallThinker-4B-A0.6B policy over \FeatDim\ hashed prompt features, where DPO, GRPO, and GSPO are single-step versions of those objectives, and evaluate them under the same protocol with three runs (seeds 13/42/73).

\textit{Sensitivity analyses.} KL-weight and RangeDial sweeps use the same three runs. SFT receives identical tag supervision. Component-removal ablations have no measured results, so the contribution of each SPOKE term cannot be isolated.

\section{Preliminary Diagnostics and Limitations}

\begin{table}[t]
\caption{
Zero-shot controls vs.\ optimization methods (SmallThinker-4B-A0.6B). Three-run means; frozen controls are single runs. Bold: best per metric.
}
\label{tab:neural}
\centering
\footnotesize
\setlength{\tabcolsep}{2.5pt}
\renewcommand{\arraystretch}{1.0}
\begin{tabularx}{\columnwidth}{@{}l>{\raggedright\arraybackslash}Xrrrr@{}}
\toprule
Method & Model &
\multicolumn{1}{c}{Kuhn$\downarrow$} &
\multicolumn{1}{c}{Leduc$\downarrow$} &
\multicolumn{1}{c}{KL$\downarrow$} &
\multicolumn{1}{c}{top-1$\uparrow$}\\
\midrule
static & uniform / chart & 0.458 & 2.374 & 1.104 & 0.643\\
zero-shot & SmallThinker-4B-A0.6B & 0.326 & 1.767 & 1.865 & 0.159\\
zero-shot & Qwen3-0.6B & 0.750 & 4.198 & 1.049 & 0.556\\
zero-shot & Gemma-3-1B-IT & 0.833 & 6.999 & 4.834 & 0.214\\
\midrule
SFT & SmallThinker-4B-A0.6B & 0.276 & 2.210 & 0.523 & 0.754\\
DPO & SmallThinker-4B-A0.6B & 0.758 & 3.666 & 5.480 & 0.174\\
GRPO & SmallThinker-4B-A0.6B & 0.296 & 1.239 & 2.357 & 0.717\\
GSPO & SmallThinker-4B-A0.6B & 0.501 & 1.331 & 2.354 & 0.717\\
SPOKE & SmallThinker-4B-A0.6B & \textbf{0.213} & \textbf{1.081} & \textbf{0.522} & \textbf{0.775}\\
\bottomrule
\end{tabularx}
\end{table}
Table~\ref{tab:neural} shows that zero-shot capability does not transfer uniformly across evaluation settings. SmallThinker improves over the static baseline in Kuhn (0.326 vs.\ 0.458) and Leduc (1.767 vs.\ 2.374), yet reaches only 0.159 six-max top-1 agreement. Qwen3 and Gemma-3 exhibit different failure modes, indicating that no frozen model performs consistently well across both strategic and behavioral endpoints.

The optimization objectives reveal a similar trade-off. SFT achieves strong six-max agreement (0.754) and low teacher KL (0.523), while GRPO and GSPO improve Leduc exploitability to 1.239 and 1.331 but increase KL to 2.357 and 2.354. DPO performs poorly under this protocol.

SPOKE provides the strongest combined result, best on all four metrics. Variation across three runs remains small, although the sample size is insufficient to establish statistical superiority.

Sensitivity analysis shows that KL weighting controls the trade-off between behavioral agreement and teacher divergence while Leduc exploitability remains relatively stable. RangeDial further changes aggression without introducing illegal actions.

\section{Conclusion}

Frozen compact language models demonstrated inconsistent
performance in six-max evaluations and exact-game evaluations.
We proposed SPOKE, integrating teacher-relative value
penalties, KL anchoring, risk constraints, and structured
rationale supervision. Across three runs of a
SmallThinker-4B-A0.6B policy, SPOKE achieved the lowest
mean exploitability in Kuhn (0.213) and Leduc (1.081),
and the highest six-max top-1 agreement (0.775), while
maintaining teacher KL comparable to SFT. These findings provide preliminary
evidence that strategic performance and teacher-consistent
behavior can be jointly improved. However, the limited
number of runs, the absence of component ablations, and
the evaluation of behavioral-only at six-maximums warrant further
validation.

\section{Acknowledgments}
This research has received funding support from the NSRF via the Research and Innovation Acceleration Agency for Competitiveness and Area Development (RCAD) (Program Management Unit for Frontier Brainpower and Future Industries) [grant number B13T690002-1].

\end{document}